\documentclass{article}

\usepackage[nonatbib, final]{neurips_2020}

\usepackage[utf8]{inputenc} 
\usepackage[T1]{fontenc}    
\usepackage{hyperref}       
\usepackage{url}            
\usepackage{booktabs}       
\usepackage{amsfonts}       
\usepackage{nicefrac}       
\usepackage{microtype}      

\usepackage{mathrsfs}
\usepackage{amsmath, amssymb, bm}
\usepackage[sort, numbers]{natbib}
\usepackage{algorithm, algpseudocode}
\usepackage{graphicx, subfigure}
\usepackage[export]{adjustbox}

\newtheorem{defn}{Definition}

\newcommand{\braces}[1]{\left\lbrace #1 \right\rbrace}
\newcommand{\parens}[1]{\left( #1 \right)}

\newcommand{\phat}{\hat{p}}

\newcommand{\ptrain}{\hat{p}_{train}}
\newcommand{\ptest}{\hat{p}_{test}}

\newcommand{\fis}[1]{f_{is}\left( #1\right)}
\newcommand{\ffid}[2]{f_{fid}\left( #1, #2 \right)}
\newcommand{\fnnd}[2]{f_{nnd}\left( #1, #2 \right)}

\newcommand{\comp}{\text{COMP}}
\newcommand{\len}{\text{LEN}}
\newcommand{\scomp}{\text{SCOMP}}
\newcommand{\const}{\mathrm{const}}
\newcommand{\norm}[1]{\left\lVert #1 \right\rVert}
\newcommand{\abs}[1]{\left\lvert #1 \right\rvert}

\newcommand{\dtrain}{\mathscr{D}_{train}}
\newcommand{\dtest}{\mathscr{D}_{test}}
\newcommand{\dr}{\mathscr{D}_r}
\newcommand{\dg}{\mathscr{D}_g}
\newcommand{\data}{\mathscr{D}}
\newcommand{\gen}{\mathcal{G}}

\newcommand{\preal}{p_{data}}
\newcommand{\pfake}{p_{model}}

\newcommand{\R}{\mathbb{R}}

\newcommand{\I}{\textbf{I}}
\newcommand{\E}[2]{\mathbb{E}_{#1} \left[ #2 \right]}
\newcommand{\KL}[2]{\text{KL} \left( #1 \vert \vert #2 \right)}

\usepackage{xcolor}

\graphicspath{{./figs}{./figs1/}}

\title{Toward a Generalization Metric for Deep Generative Models}

\author{%
  Hoang Thanh-Tung \\
  Deakin University \\
  \texttt{hoangtha@deakin.edu.au} \\
  \And
   Truyen Tran \\
   Deakin University \\
   \texttt{truyen.tran@deakin.edu.au} \\
}

\begin{document}

\maketitle

\begin{abstract}
Measuring the generalization capacity of Deep Generative Models (DGMs) is difficult because of the curse of dimensionality.
Evaluation metrics for DGMs such as Inception Score, Fréchet Inception Distance, Precision-Recall, and Neural Net Divergence try to estimate the distance between the generated distribution and the target distribution using a polynomial number of samples.
These metrics are the target of researchers when designing new models.
Despite the claims, it is still unclear how well can they measure the generalization capacity of a generative model.
In this paper, we investigate the capacity of these metrics in measuring the generalization capacity.
We introduce a framework for comparing the robustness of evaluation metrics.
We show that better scores in these metrics do not imply better generalization.
They can be fooled easily by a generator that memorizes a small subset of the training set.
We propose a fix to the NND metric to make it more robust to noise in the generated data.
Toward building a robust metric for generalization, we propose to apply the Minimum Description Length principle to the problem of evaluating DGMs.
We develop an efficient method for estimating the complexity of Generative Latent Variable Models (GLVMs).
Experimental results show that our metric can effectively detect training set memorization and distinguish GLVMs of different generalization capacities.
\end{abstract}

\section{Introduction}

The quality of a generative model is defined by the distance between the model distribution and the target distribution.
Traditional distances/divergences such as the KL divergence or the Wasserstein distance are intractable because of their exponential sample complexities.
In the last few years, a number of evaluation metrics with polynomial sample complexities have been introduced.
Some of the most common metrics are the Inception Score (IS) \citep{improvedgan}, Fréchet Inception Distance (FID) \citep{ttur}, Precision-Recall metrics \citep{precisionRecallKmeans, precisionRecallKnn, howGoodGAN}.
These metrics are intuitive and fast to compute, but some researchers have raised concerns regarding their reliability \citep{towardGenMetricGAN}.
\cite{towardGenMetricGAN} proposed to use the Neural Net Divergence (NND) \citep{equiAndGeneralization} to measure the generalization capacity of generative models.
In this paper, we evaluate the ability to measure generalization of the above evaluation metrics.
We find that all of them cannot reliably measure generalization.
They can be fooled easily by a generative model that memorizes a small subset of the training set.

We propose a novel generalization metric for a broad class of generative models called the Generative Latent Variable Models.
Our metric is based on the Minimum Description Length principle \citep{kolmogorov1968three, rissanenMDL, stochasticComplexity86, stochasticComplexity87, refinedMDL}.
The metric combines the complexity of the generative model and the divergence between the model distribution and the target distribution into a single value.
The metric inherits the robustness of the MDL framework and can be computed in polynomial time using a polynomial number of samples.

\textbf{Notations}\\
\begin{table}[!ht]
\begin{tabular}{l}

$\bm z \in \R^{d_z}$: a latent variable. \\
$\bm x \in \R^{d}$: a data point. \\
$\gen(\cdot; \bm \theta)$: a generative model with parameter $\bm \theta \in \R^{d_\theta}$.  \\
$\dr = \braces{\bm x_1, ..., \bm x_k \ \vert\ \bm x_i \sim \preal}$: a set of samples from $\preal$.  \\
$\dg = \braces{\bm x_1, ..., \bm x_l \ \vert\ \bm x_i \sim \pfake}$: a set of samples from $\pfake$. \\
$\dtrain =  \braces{\bm x_1, ..., \bm x_n \ \vert\ \bm x_i \sim \preal}$: the training set, $|\dtrain| = \textnormal{poly}(d)$. \\
$\dtest  =  \braces{\bm x_1, ..., \bm x_m \ \vert\ \bm x_i \sim \preal,\ \bm x_i \notin \dtrain}$: the test set, $|\dtest| = \textnormal{poly}(d)$. \\
$\phat_{\data}$: the uniform distribution over $\data$.  \\
\end{tabular}
\end{table}

\section{Background and Related Work}

\subsection{Divergence based evaluation metrics}
\label{sec:relatedDiv}
A generative model $\gen$ is trained on $\dtrain$ to produce $\pfake$, an estimate of $\preal$.
The quality of $\gen$ is defined by the divergence/distance between $\preal$ and $\pfake$.
Let $f$ be a divergence/distance function. 
Because $f(\preal, \pfake)$ is intractable for most distributions of interest, we approximate it with $f(\dr, \dg)$ where ${\dr}, {\dg}$ are polynomial sized datasets.

\textbf{Inception Score} (IS) \citep{improvedgan} is defined as: 
\begin{align}
\textnormal{IS}\parens{\pfake} = \exp\parens{\KL{p(Y | \bm X)}{p(Y)}}
\end{align}
where $\bm x \sim \pfake$ is the input, $y$ is the label,  and $p(y) = \E{\bm x \sim \pfake}{p(y | \bm x)}$. 
$p(y | \bm x)$ is computed by feeding the generated data point through a pretrained classifier, e.g. the Inception network \citep{inceptionModel}. 
IS is maximized when $\textnormal{H}\parens{Y | \bm X} = 0$ and $\textnormal{H}(Y) = \ln C$ where $C$ is the number of classes.
\citet{noteIS} showed that $\textnormal{IS}\parens{\pfake} \le C$.
To make our writing consistent, we convert all of the metrics in this paper to (pseudo) divergences. 
A lower divergence should imply that $\pfake$ is closer to $\preal$.
We define the following pseudo divergence:
\begin{equation}
\fis{\pfake} = C - \exp\parens{\KL{p(Y | \bm X)}{p(Y)}}
\end{equation}
This is a pseudo divergence because (1) it only contains 1 distribution, $\pfake$, the other distribution is implicitly the dataset on which the classifier was trained,  and (2) $\fis{\pfake} = 0$ does not imply that the two distributions are the same (more on this in Section~\ref{sec:capacity}). 
A fundamental flaw of IS is that a smaller $\fis{\pfake}$ implies that $p(y)$ is closer to the uniform distribution, not that $p(y)$ is closer to the true distribution over the labels.
Fortunately (or unfortunately), the distribution over the labels in the Imagenet dataset \citep{imagenet} is uniform.
This makes IS a popular and somewhat abused tool for evaluating image generation models.

\textbf{Fréchet Inception Distance} (FID): \citet{ttur} approximate $\preal$, $\pfake$ with two Gaussian distributions $\mathcal{N}\parens{\bm \mu_r, \bm \Sigma_r}$ and $\mathcal{N}\parens{\bm \mu_g, \bm \Sigma_g}$, respectively, and then compute the Fréchet distance between the two Gaussians:
\begin{equation}
\ffid{\preal}{\pfake} = \norm{\bm \mu_r - \bm \mu_g}_2^2 + \textnormal{Tr}\left(\bm \Sigma_r + \bm \Sigma_g - 2 \left( \bm \Sigma_r\bm \Sigma_g \right)^{\frac{1}{2}} \right)
\end{equation}
Because FID compares only the first 2 moments of $\preal$ and $\pfake$, $\ffid{\preal}{\pfake} = 0$ does not guarantee that the 2 distributions are the same.

\textbf{Neural Net Divergence} (NND) \citep{towardGenMetricGAN, equiAndGeneralization} uses neural nets to compute the divergence between $\preal$ and $\pfake$
\begin{equation}
\label{eqn:nnd}
\fnnd{\preal}{\pfake} = \underset{f_{\bm \phi} \in \mathcal{F}}{\sup} \parens{ \E{\bm x \sim \preal}{f_{\bm \phi}(\bm x)} - \E{\bm x \sim \pfake}{f_{\bm \phi}(\bm x)} }
\end{equation}
where $f_{\bm \phi}: \R^{d} \rightarrow \R$ is usually a neural network with finite capacity.
$f_{\bm \phi}$ is trained to maximize the difference in Eqn.~\ref{eqn:nnd}.
\citet{equiAndGeneralization} showed that NND can detect the lack of diversity in generated data.
\citet{towardGenMetricGAN} leveraged that capability to estimate the generalizability of GMs.
We show in Section~\ref{sec:capacity} that the diversity in generated data can be faked easily by adding white noise to the data.
This fake diversity is present in real-world models such as GANs and VAEs.
We propose a fix to NND that removes the effect of fake diversity while retaining the discriminative power of NND. 

\subsection{Precision-Recall based evaluation metrics}
\label{sec:relatedPR}
The above metrics use scalars to measure both the quality and diversity of generated samples.
The following metrics are designed to separate quality from diversity.

\textbf{$k$-means based Precision-Recall} ($k$-means PR) \citep{precisionRecallKmeans} applies $k$-means algorithm to $\dr \cup \dg$, builds 2 histograms to approximate $\preal$ and $\pfake$, and computes the difference between these histograms. 
The results are 2 F-scores, $F_\beta$ and $F_{1/\beta}$.
For $\beta > 1$, $F_\beta$ can be interpreted as the recall (diversity) and $F_{1/\beta}$ can be interpreted as the precision (quality) of the generated data (Algo.~\ref{algo:kmeans}).
We define the following pseudo divergences: $f_{p\textnormal{-}kmeans}= 1 - F_{1/\beta}$, $f_{r\textnormal{-}kmeans}= 1 - F_{\beta}$.

\begin{figure}
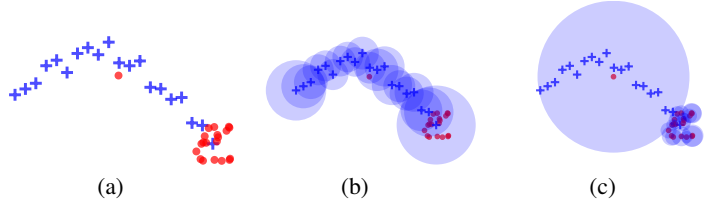

\centering
\setlength{\fboxsep}{1pt}%
\setlength{\fboxrule}{1pt}%
\subfigure[]{
\adjincludegraphics[width=0.22\textwidth, trim={{0.42\width} {0.17\width} {0.246\width} {0.3\width}}, clip]{figs/knn_data.pdf}\label{fig:knnData}}
\subfigure[]{
{\adjincludegraphics[width=0.22\textwidth, trim={{0.34\width} {0.12\width} {0.19\width} {0.211\width}}, clip]{figs/knn_real.pdf}\label{fig:knnReal}}}
\subfigure[]{
{\adjincludegraphics[width=0.22\textwidth, trim={{0.34\width} {0.12\width} {0.19\width} {0.211\width}}, clip]{figs/knn_fake.pdf}\label{fig:knnFake}} }
\caption[$k$-nn PR on a 2-dimensional dataset.]{$k$-nn PR on a 2-dimensional dataset. \subref{fig:knnData} real data (blue crosses) and fake data (red dots). \subref{fig:knnReal}, \subref{fig:knnFake} real and fake manifolds approximated with $3$-nearest neighbors.
Although the real and the fake data are very different, the precision and recall of the fake data are 1 and 1.}
\label{fig:knn}
\end{figure}

\textbf{$k$-nn based Precision-Recall} ($k$-nn PR) \citep{precisionRecallKnn} computes the precision and recall by comparing the reconstructed real manifold and the reconstructed fake manifold (Algo.~\ref{algo:knn} and Fig.~\ref{fig:knn}).
\citet{precisionRecallKnn} claimed that $k$-nn PR has better discriminative power than $k$-means PR and demonstrated that on several datasets.
However, we show in Section~\ref{sec:capacity} that $k$-nn PR has the worst worst-case performance, and in practice, it cannot distinguish a good model from a very bad model.
We define the following pseudo divergences: $f_{p\text{-}knn} = 1 - P$, $f_{r\text{-}knn} = 1 - R$.

Similar to IS and FID, $k$-means PR and $k$-nn PR assign (near) perfect scores to generative models that produce the entire training set.

\begin{minipage}{0.7\textwidth}
\begin{algorithm}[H]
    \centering
    \caption{$k$-means PR} \label{algo:kmeans}
    \begin{algorithmic}[1]
        \State \textbf{Inputs}: $\data_r,\ \data_g$, hyper parameter $\beta$
        \State \textbf{Outputs}: $ F_\beta,\ F_{1/\beta}$
        \State $\mathscr{C} \gets kmeans(\data_r \cup \data_g)$
        \State $hist_g[i] \gets \frac{\text{\# fake data points in cluster } \mathscr{C}_i}{|\data_g|}$
        \State $hist_r[i] \gets \frac{\text{\# real data points in cluster } \mathscr{C}_i}{|\data_r|}$
        \State $curve = pr\_curve(hist_r, hist_g)$
        \State ${F_\beta = f\_score(curve, \beta)}$
        \State ${F_{1/\beta} = f\_score(curve, 1/\beta)}$
        \State \Return $ F_\beta,\ F_{1/\beta}$
    \end{algorithmic}
\end{algorithm}
\end{minipage}

\begin{minipage}{0.7\textwidth}
\begin{algorithm}[H]
    \centering
    \caption{$k$-nn PR} \label{algo:knn}
    \begin{algorithmic}[1]
        \State \textbf{Inputs}: $\data_r,\ \data_g$
        \State \textbf{Outputs}: Precision $P$, Recall $R$
        \State $\mathcal{M}_g \gets \emptyset$ \quad // build the fake manifold
        \For{\text{each data point } $\bm x_{i} \in \data_g$}
        \State $\bm x_j \gets $ the $k$-th nearest neighbor of $\bm x_i$
        \State $S_i \gets sphere(c=\bm x_i, r=\norm{\bm x_i - \bm x_j})$
        \State $\mathcal{M}_g \gets \mathcal{M}_g \cup S_i$
        \EndFor
        \State \text{build the real manifold } $\mathcal{M}_r$
        \State $P \gets \frac{\text{\# fake data points in } \mathcal{M}_r}{|\data_g|}$
        \State $R \gets \frac{\text{\# real data points in } \mathcal{M}_g}{|\data_r|}$
        \State \Return $P,\ R$
    \end{algorithmic}
\end{algorithm}
\end{minipage}

\subsection{Other evaluation metrics}

\textbf{Metrics for class-conditional models}: \citet{realTimeSeriesTSTR} proposed the ``Train on Synthetic Test on Real'' (TSTR) approach to evaluate class-conditional GANs.
The metric is expensive to compute as it requires training a classifier on a labeled synthetic dataset.
\citet{howGoodGAN} extended the idea to a Precision-Recall-like metric for class-conditional GANs.
Because these metrics are applicable only to class-conditional GMs, we do not consider them in this paper.

\textbf{Topological/Geometrical approaches}: \citet{geometryScore, topologyDistance} applied geometrical and topological tools to measure the difference between the fake manifold and the real manifold.
These metrics are not designed for measuring generalizability and are not considered in this paper.
\citet{styleGAN} used the \textit{pairwise perceptual distance} - the pairwise distance between data points on the feature manifold - to measure the smoothness of the feature manifold.
The intuition is that if the perceptual distance is small, then the manifold is smoother, and the GM might learn better representations and have better generalizability. 
We show in Section~\ref{sec:mdl} that the pairwise distance does not well reflect the smoothness of the manifold, and it is minimized when total mode collapse occurs.
Therefore, pairwise perceptual distance is not a good metric for GMs.

\textbf{Non-parametric approaches}:
\citet{doGANLearnDist} used the birthday paradox and human evaluation to estimate the number of distinct modes that a GM can generate. 
The method cannot be automated and has high variance and bias because humans are involved in the loop. 
\citet{nonParametricTestDataCopying} proposed a three sample test for data copying in GMs.
The metric can detect data copying but cannot measure the generalizability of a model or the quality of the generated samples.

\subsection{A generative model with poor generalization capacity}
\label{sec:poorG}
When mode collapse or training set memorization \citep{gan} happen, the generative model generates slightly different versions of the memorized data points. 
We simulate this kind of behavior with the following generative procedure: 

Given a dataset $\data \subseteq \dtrain$, a constant $\epsilon \in \R$, a new data point is generated by

\begin{enumerate}
\item Draw a data point $\bm x \in \data$ at random
\item Draw a noise vector $\bm u \sim p_u$, where $p_u$ is the noise distribution.
\item Output the noisy data point: $\tilde{\bm x} = \bm x + \epsilon \bm u$
\end{enumerate}

In our experiments, $p_u = \mathcal{U}(-\bm 1, \bm1)$ (see Fig.~\ref{figx:noisyImages} for samples from MNIST dataset).
We tried replacing $\mathcal{U}(-\bm 1, \bm1)$ with $\mathcal{N}(\bm 0, \I)$ and found no difference in performance.
This generator, denoted as $\gen_{\data,\epsilon}$, has no real generalization capacity and the number of essentially different modes in $p_{\gen_{\data,\epsilon}}$ is $\abs{\data}$.
For $\epsilon > 0$, this generator can be approximated with a neural net with finite capacity.
\section{Evaluating evaluation metrics}
\label{sec:capacity}

Given a training set $\dtrain \sim \preal$, we train a generative model $\gen$ on $\dtrain$ and get a distribution $\pfake$.
We would like to estimate the divergence/distance between $\preal$ and $\pfake$ using a divergence/distance function $f$.
Because the true divergence/distance $f(\preal, \pfake)$ is intractable, we approximate it with the empirical divergence/distance $f(\dtest, \dg)$ where $\dtest \sim \preal$ and $\dg \sim \pfake$.
If $\dtest = \dtrain$ then $\gen$ can achieve the perfect score by memorizing the entire $\dtrain$.
To reduce the effect of training set memorization, we require that $\dtest \cap \dtrain = \emptyset$.
We want  $f(\dtest, \dg)$ to track $f(\preal, \pfake)$, i.e.~$f(\dtest, \dg)$ is small only when $f(\preal, \pfake)$ is small.
Specifically, we want $f$ to satisfy:
\begin{align}
\E{\dtest,\ \data_1 \sim \preal,\ \abs{\dtest} = m, \ \abs{\data_1} = n_1}{f(\dtest, \data_1)} > \E{\dtest,\ \data_2 \sim \preal,\ \abs{\dtest} = m, \ \abs{\data_2} = n_2}{f(\dtest, \data_2)} \nonumber \\ 
\text{ for all } n_1 < n_2
\label{eqn:requirement}
\end{align}
Intuitively, the divergence/distance should decrease as the generative model can generate more (significantly different) data points from the target distribution.
We say that a generative model generalizes if it can generate more data points from the target distribution than a generative model that memorizes the training dataset.
To test for generalization, we generate a dataset $\dg$ whose size is larger than that of $\dtrain$ and compare $f(\dtest, \dg)$ and $f(\dtest,\dtrain)$.

We show in the next subsections that Inception Score, Precision-Recall, and Neural Net Divergence do not satisfy Eqn.~\ref{eqn:requirement} and are not robust against training set memorization.

\subsection{Worst-case analyses of evaluation metrics}
Let $\data^*$ be the smallest subset of $\dtrain$ that satisfies $\E{\dtest \sim \preal,\ \abs{\dtest} = m}{f(\dtest, \data^*)} \le \E{\dtest \sim \preal,\ \abs{\dtest} = m}{f(\dtest, \dtrain)}$.
The divergence $f$ cannot distinguish a model that produces only $\data^*$ and a model that produces $\dtrain$.
By memorizing only $\data^*$, a generator can fool $f$ that it can produce a larger dataset.
The smaller $\abs{\data^*}$ is, the more vulnerable $f$ is to training set memorization.
We construct $\data^*$ for the metrics mentioned above and discuss the relationship between $\data^*$ and the performance of a metric in practice.
Because NND is the only metric designed to be robust against training set memorization, it is the focus of this section. 
We defer the details for other metrics to the appendix.

\textbf{Inception Score}  
\begin{equation}
\data^*_{is} = \braces{\bm x_1, ..., \bm x_C \ \vert \ \bm x_i \in \mathscr{C}_i}
\label{eqn:isD}
\end{equation}
where $\mathscr{C}_i \subset \dtrain$ is the $i$-th class of the training set. Details in Appx.~\ref{appx:is}.

\textbf{$k$-means Precision-Recall}
\begin{equation}
\data^*_{kmeans} = \braces{\bm x_1, ..., \bm x_k \ \vert\ \bm x_i \in \mathscr{C}_i}
\label{eqn:kmeansD}
\end{equation}
where $\mathscr{C}_i$ is the $i$-th cluster of the $k$-means clustering algorithm. Details in Appx.~\ref{appx:kmeans}.

\textbf{$k$-nn Precision-Recall}
\begin{equation}
\data^*_{knn} = \braces{\bm x_1,\ \bm x_2 \ \vert\ \bm x_1, \bm x_2 \in \dtrain \text{ and } \norm{\bm x_1 - \bm x_2} \ge \norm{\bm x_i - \bm x_j} \forall \bm x_i, \bm x_j \in \dtrain}
\label{eqn:knnD}
\end{equation}
Details in Appx.~\ref{appx:knn}.

\textbf{Neural Net Divergence} is computed by training a finite capacity neural net $f_{\bm \phi}$ to maximize the objective in Eqn.~\ref{eqn:nnd}. 
Because the value of NND depends on the network and the training procedure, there is no fixed $\data^*_{nnd}$.
However, given a network $f_{\bm \phi}$, we can still estimate the size of $\data^*_{nnd}$ through experiments. 

\begin{figure}
\centering
\subfigure[]{\includegraphics[width=0.32\textwidth]{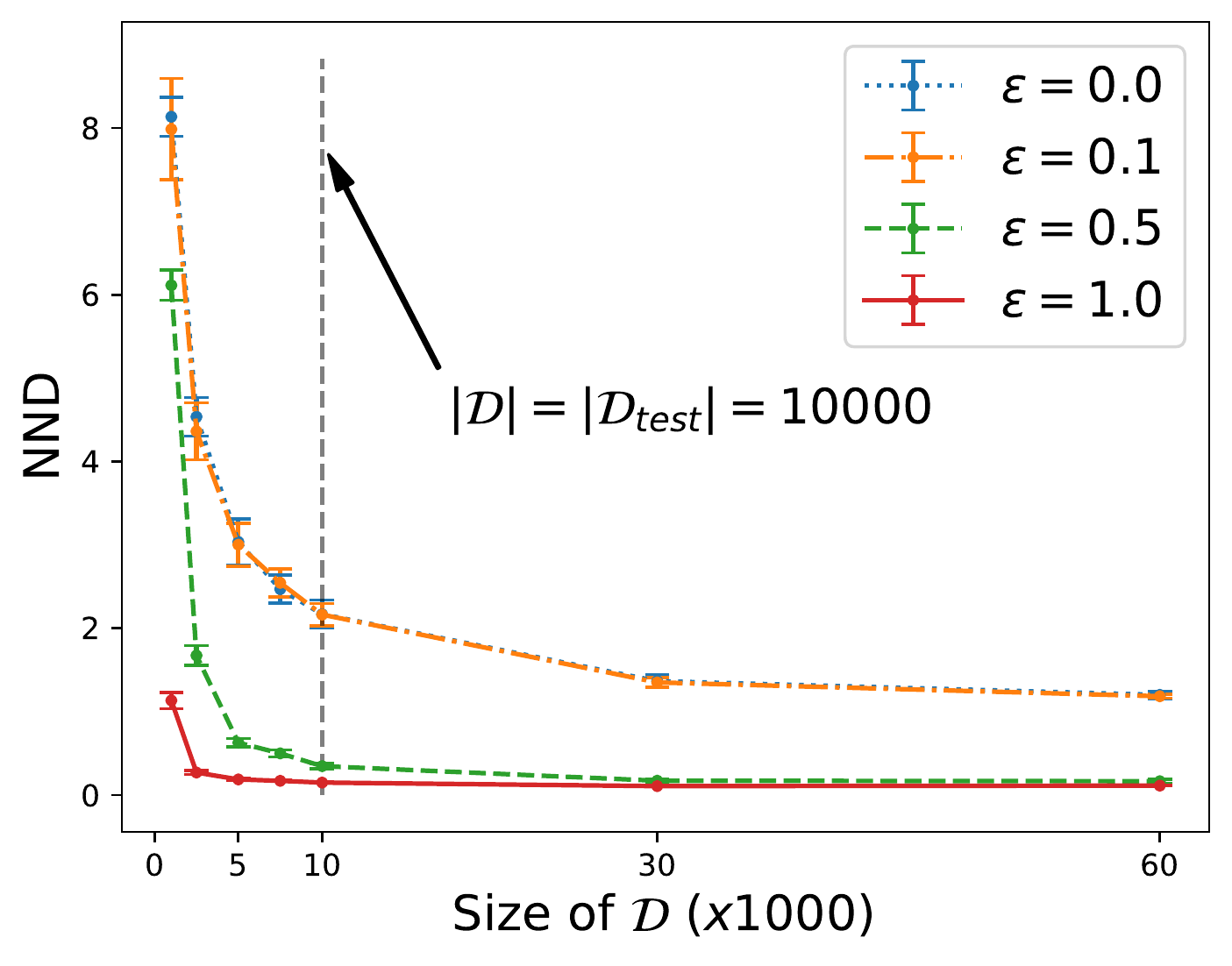} \label{fig:noiseInf}}
\subfigure[]{\includegraphics[width=0.32\textwidth]{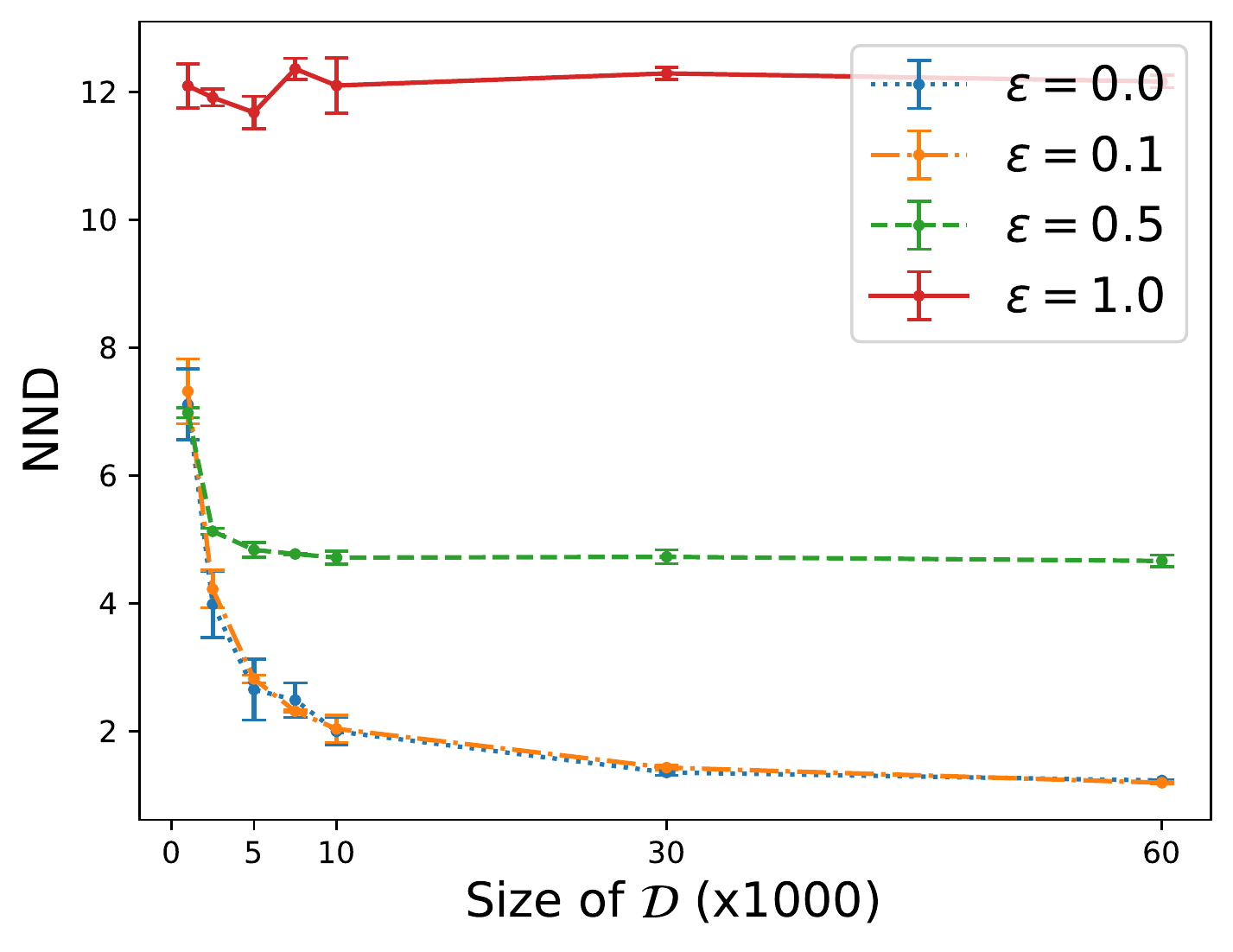} \label{fig:noiseFixed}}
\subfigure[]{\includegraphics[width=0.32\textwidth]{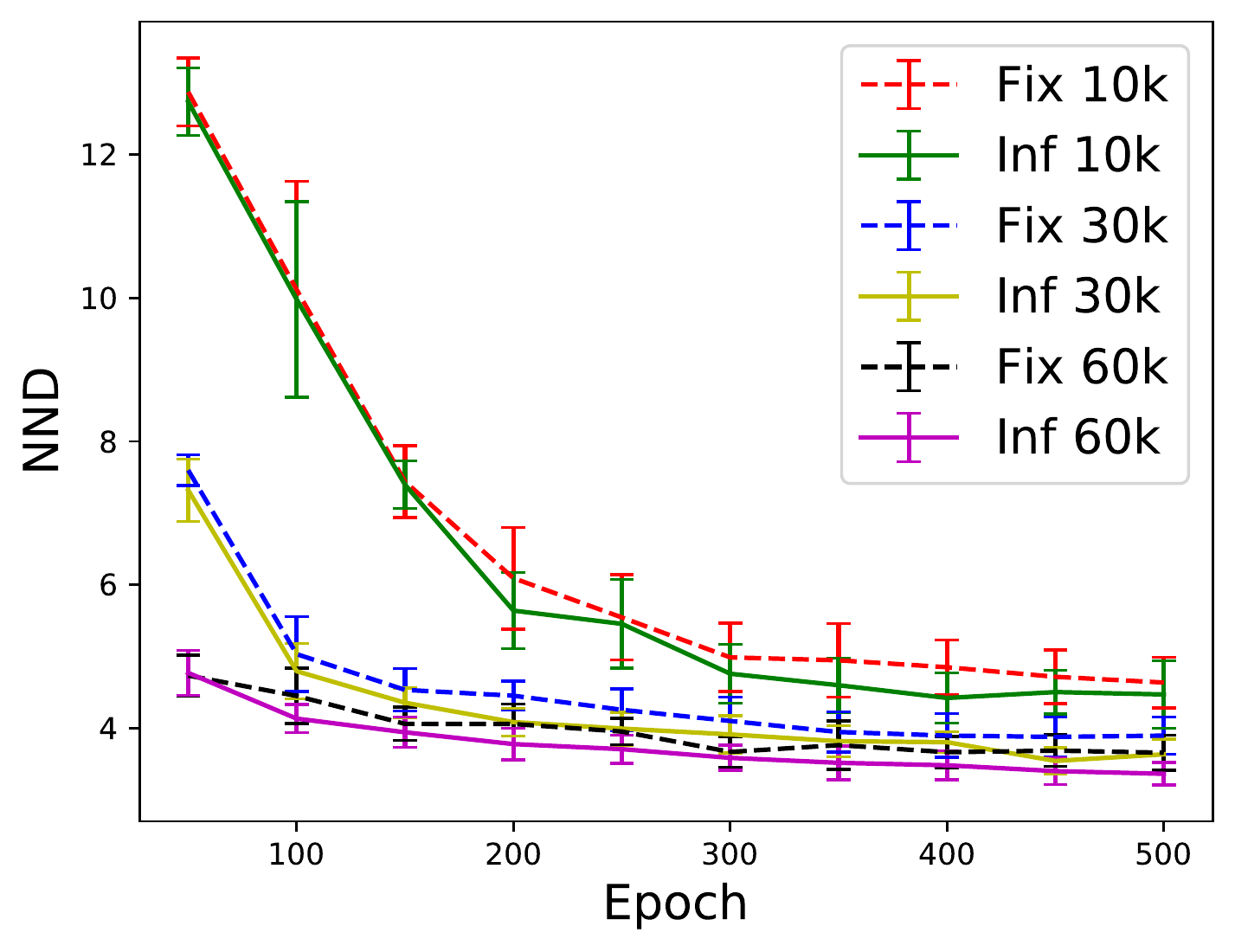}\label{fig:infVSFixed}}
\caption[The effect of noise on NND]{The effect of noise on NND. 
\subref{fig:noiseInf}: noisy samples were generated continuously throughout training. 
\subref{fig:noiseFixed}: a fixed number of noisy samples were used in training.
\subref{fig:infVSFixed}: continuously generated fake data (labeled `Inf + training set size') v.s. fixed size fake data (labeled `Fix + training set size').}
\label{fig:nndBreak}
\end{figure}

First, we investigate the effect of noise on NND. 
In our experiment, the dataset is the MNIST dataset, $f_{\bm \phi}$ is an MLP with 3 hidden layers and 512 neurons in each layer. 
Details about the architecture and hyper parameters are in Appx.~\ref{appx:nnd}.
We use the generator described in Section~\ref{sec:poorG} to generate noisy versions of data points in $\data$, a random subset of $\dtrain$ (see Fig.~\ref{figx:noisyImages} for noisy images generated by the generator).
The number of essentially different data points that the generator can generate is $\abs{\data}$.
We vary $\abs{\data}$ from $1000$ to $60000$ and $\epsilon$ from $0$ to $1$.
The noisy samples are generated continuously throughout the training process as recommended by the authors NND metric, \cite{towardGenMetricGAN}.
As we can see in Fig.~\ref{figx:noisyImages}, for $\epsilon = 0.1$ and $\epsilon = 0.5$ the noisy images are very clear and easily recognizable.
For $\epsilon = 1$, the images are much more noisy but still recognizable.
At first glance, one might think that adding noise to clean data will increase the NND because the quality of generated data is worsened.
The result in Fig.~\ref{fig:noiseInf} shows the opposite.
As we can see from the figure, when $\epsilon$ and $\abs{\data}$ increase, $f_{nnd}$ decreases.
Increasing $\epsilon$ results in a bigger decrease in $f_{nnd}$ than increasing $\abs{\data}$. 
For $\epsilon = 1$, a noisy random subset of size 1000 outperforms a noiseless set of size 60000 by a large margin.
The network also cannot distinguish noisy sets of size 10000 and 60000 and gives them the same score (the red solid line in Fig \ref{fig:noiseInf}).
We conclude that $\abs{\data^*_{nnd}} \le 1000$ for this specific network.\footnote{We note that our experiments with $p_u = \mathcal{N}(\bm 0, \I)$ produces the same results.}

On the positive side, we observe that the NND decreases as $\abs{\data}$ increases.
However, the NND is plateauing as $\abs{\data}$ reaches $60000$.
If we continue to increase $\abs{\data}$ to much larger numbers, the network will assign about the same NND to these datasets.
This behavior is expected as the capacity of the network is polynomial in $d$.
A neural network of a polynomial size can only distinguish datasets of polynomial sizes (for a more detailed discussion, please refer to \cite{equiAndGeneralization}).
We can still use NND to test for generalization if the network can distinguish datasets of sizes larger than $\abs{\dtrain}$.
The larger the network is, the more discriminative the NND becomes.

If we can remove the fake diversity caused by noise,  we can use NND to estimate the diversity of generated data.
We can reduce the effect of noise by generating a noisy data set $\dg$ of a fixed and polynomial size and train $f_{\bm \phi}$ on $\dg$ and $\dtest$.
For the experiment in Fig.~\ref{fig:noiseFixed}, we use $\gen_{\data, \epsilon}$ to generate a dataset $\dg$ of $60000$ noisy data points.
In contrast to the previous experiment, the NND increases with the level of noise.
This behavior is expected as a larger $\epsilon$ results in worse images, making the task of separating $\dg$ from $\dtest$ easier.
By fixing the size of $\dg$, we can effectively remove the fake diversity.

The outputs of GMs usually contain noise.
For Autoregressive Models and Energy Based Models, the noise comes from the randomness in the sampling process.
For generative latent variable models like GANs, VAEs, and flows, the noise comes from the prior distribution.
A generative model can memorize a subset of the training data, add noise to the memorized data points, and fool $f_{nnd}$ that it has learned to faithfully reproduce the target distribution.
Our next experiment investigates this phenomenon in real-world models. 
We use GANs as our example.
We select a random subset $\data$ from $\dtrain$,  train a GAN on $\data$, and use the resulting model to generate fake data.
For each model, we compute 2 NND scores following the original procedure and our new procedure described in the previous paragraph.
Fig.~\ref{fig:infVSFixed} shows the changes in NND of models over 500 epochs.
As expected, the NND of `Fix' datasets is higher than that of `Inf' datasets.
However, the difference between `Fix' and `Inf' datasets is small, suggesting that the level of noise in the outputs of GAN is much lower than that in our experiment in Fig.~\ref{fig:noiseInf}.

\section{An MDL inspired Generalization Metric}
\label{sec:mdl}

In this section, we present a novel metric for measuring the generalization capacity of generative latent variable models.
The metric aims to fix the problems of previously mentioned metrics.
Our metric is based on the manifold hypothesis and the Minimum Description Length principle.
We assume that the data lie on a high-dimensional manifold and there is a path between every pair of points on the manifold.

\subsection{Motivation}
\label{sec:motivation}

\begin{figure}
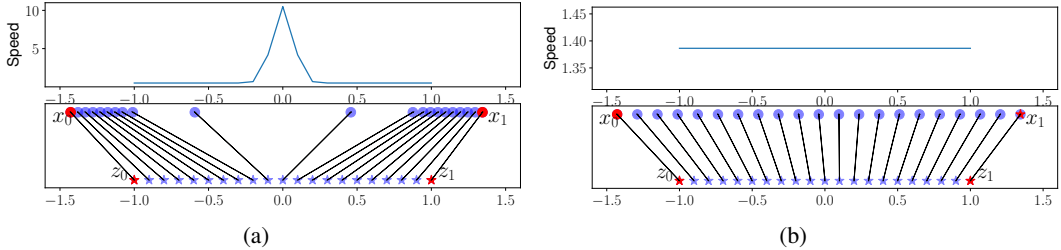

\subfigure[]{
\adjincludegraphics[width=0.5\textwidth]{figs/interExplode.pdf}\label{fig:interExplode}}
\subfigure[]{
\adjincludegraphics[width=0.5\textwidth]{figs/interConstant.pdf}\label{fig:interConstant}}

\caption[Detecting memorization using interpolation speed]{The target distribution is the uniform distribution over the line segment $(\bm x_0, \bm x_1)$. \subref{fig:interExplode} a GM that memorizes the training set.
\subref{fig:interConstant} a GM that produces constant speed interpolation.}
\label{figx:inter}
\end{figure}

\subsubsection{Preliminaries}
\textbf{Path length and speed}: A path from $\bm z_0$ to $\bm z_1$ is a continuous function $\bm z(t): [0, 1] \rightarrow \mathbb{R}^{d_z}$ that 
satisfies $\bm z(0) = \bm z_0,\ \bm z(1) = \bm z_1$. 
The Frobenius norm of the Jacobian $\frac{\partial \bm z}{\partial t}$ is the speed of the path at time $t$.
The path has a constant speed if $\norm{\frac{\partial \bm z}{\partial t}}_F = \const$. 
Let $\gen$ be a continuous function from $\R^{d_z}$ to $\R^d$.
$\gen$ maps a path from $\bm z_0$ to $\bm z_1$ to a path from $\bm x_0 = \gen(\bm z_0)$ to $\bm x_1 = \gen(\bm z_1)$.
The length of the path from $\bm x_0$ to $\bm x_1$ is 
\begin{align}
\ell(\bm x_0, \bm x_1) = \int_0^1 \norm{\frac{\partial \bm x}{\partial t}}_F dt = \int_0^1 \norm{\frac{\partial \bm x}{\partial \bm z} \frac{\partial \bm z}{\partial t}}_F dt \label{eqn:pathLength} 
\end{align}

\textbf{Minimum Description Length principle }\citep{kolmogorov1968three, rissanenMDL, stochasticComplexity86, stochasticComplexity87, refinedMDL} (see \cite{mdlTut} for a comprehensive tutorial) is a formalization of the Occam's razor principle. 
Refined MDL \citep{refinedMDL} defines the stochastic complexity of a dataset $\data$ given a parametric model (a set of parametric hypotheses) $\mathcal{H}$ as  
\begin{align}
\text{SCOMP}(\data | \mathcal{H}) = \len(\data | \hat{H}) + \comp(\mathcal{H}) \label{eqn:scomp}
\end{align}
where $\len(\data | H) = -\log p(\data | H)$ is the description length of $\data$ given a point hypothesis $H \in \mathcal{H}$, $\hat{H} \in \mathcal{H}$ is the point hypothesis that maximizes the probability of $\data$, and $\comp(\mathcal{H})$ is the parametric complexity of $\mathcal{H}$. 
The MDL principle states that the model with the smallest $\scomp(\data | \mathcal{H})$ is the model with the best generalization capacity.
Let $\mathcal{H}_L$ be the set of $L$-Lipschitz functions, then the bigger $L$ is, the more \textit{essentially different} functions $\mathcal{H}_L$ contains (c.f. Section 2.6.2 in \cite{mdlTut}).
$L$ and $\comp(\mathcal{H}_L)$ are positively correlated. In the following, we use $L$ in place of the parametric complexity of $\mathcal{H}_L$.

\subsubsection{Detecting memorization and estimating complexity}
A metric for generalization must be able to detect memorization.
We study the behavior of GLVMs when they memorize training data.
Fig.~\ref{fig:interExplode} illustrates the situation where a GLVM $\gen_1$ tries to memorize a training set $\data$  (shown by two red dots). 
For any latent code $\bm z$ (blue stars), $\gen_1$ tries to make $\bm x = \gen(\bm z)$ (blue dots) as close as possible to a training data point. 
Because generated data points are concentrated in 2 clusters, when we interpolate from $\bm z_0$ to $\bm z_1$ at constant speed $\norm{\frac{\partial \bm z}{\partial t}}_F$, the speed $\norm{\frac{\partial \bm x}{\partial t}}_F$ blows up in the middle of the path (top pane in Fig.~\ref{fig:interExplode}).
Fig.~\ref{fig:interConstant} shows another GLVM $\gen_2$ that produces constant speed path from $\bm x_0$ to $\bm x_1$. 
Although the maximum speed in Fig.~\ref{fig:interExplode} is much higher than that in Fig.~\ref{fig:interConstant}, the paths in the two figures have the same length.
Compared to the path length,  the maximum speed is a better feature for detecting memorization (see Fig.~\ref{fig:mdl}).

From Eqn.~\ref{eqn:pathLength}, 
we see that if $\norm{\frac{\partial \bm z}{\partial t}}_F = \const$, then the maximum speed $s_{max}(\bm x_0, \bm x_1) = \max_{t \in [0, 1]}\left(\norm{\frac{\partial \bm x}{\partial t}}_F\right)$ is proportional to the Lipschitz constant of $\gen$ on the path. 
The parametric complexity of $\gen$ can be defined as the expectation of $s_{max}$ over the latent space  
\begin{align}
\comp(\gen) = \mathbb{E}_{\bm z_i, \bm z_j \sim p_z}[s_{max}(\gen(\bm z_i), \gen(\bm z_j))] \label{eqn:compg}
\end{align}
We use the average of $s_{max}$ instead of the absolute maximum speed because (1) the absolute max speed has too high variance, making it an uninformative quantity, and (2) the average tracks the cost of describing the generated manifold more closely than the absolute maximum speed.

\subsection{Definition}
\label{sec:metricDef}

Because $-\log(\data | H)$, the description length of the training data given the GLVM is not easily computable in some GLVMs like GANs, we replace it with the divergence between $\hat{p}_{data}$ and $\hat{p}_{model}$.
The divergence can be computed using one of the methods in Section~\ref{sec:relatedDiv}.
\begin{defn}
Let $\dtrain$ be a set of i.i.d. samples from a distribution $\preal$. The \emph{generalization metric} for a GLVM $\gen$ trained on $\dtrain$ is defined as:
\begin{align}
f_{gen}(\gen) = \alpha f(\dtrain, \dg) + \comp(\gen)
\end{align}
where $f(\dtrain, \dg)$ is the divergence between the training data and generated data, $\comp(\gen)$ is the parametric complexity in Eqn \ref{eqn:compg}, and $\alpha > 0$ is a scalar that controls the relative importance of the divergence to the complexity.
\end{defn}
If $\gen$ memorizes $\dtrain$, then $f(\dtrain, \dg)$ goes to 0 but $\comp(\gen)$ blows up.
If $\comp(\gen)$ goes to 0, then generated samples are concentrated in a small region, making $f(\dtrain, \dg)$ goes up (assuming that $f$ is a divergence that is good at detecting the lack of diversity, e.g., our fixed NND).
Therefore, our metric cannot be fooled by training set memorization.
In general, $f_{gen}$ cannot be 0 because the two terms cannot be 0 at the same time. 

A metric with large $\alpha$ will prefer models that can reproduce the training set more faithfully.
A small $\alpha$ will result in a metric that prefers simpler models.
$\alpha$ can be removed if we are comparing models with the same divergence.

\subsection{A constant speed regularizer for GANs}
A GLVM that produces constant speed paths in the space of generated data is desirable because (1) it is less likely to memorize the training set, (2) it does not make sudden jumps from one memorized data point to another, thus produces smoother interpolation.
We propose the following constant speed regularizer for the generator $\gen$ in GANs:
\begin{align}
\mathcal{L}^{\const}_\gen = \mathcal{L}_\gen + \lambda \mathbb{E}_{\bm z_i, \bm z_j \sim p_z}\left[\mathbb{E}_{t \in [0, 1]}\left[\left(\norm{ \frac{\partial \gen(\bm z(t))}{\partial t}} - \bar{s}\right)^2 \right]\right]
\end{align}
where $\mathcal{L}_{\gen}$ is the original loss function of $\gen$, $\bar{s}$ is the average speed on the path from $\gen(\bm z_i)$ to $\gen(\bm z_j),$ and the interpolation method is a constant speed interpolation method, i.e. $\norm{\frac{\partial \bm z}{\partial t}} = \const$.
The regularizer forces the variance of the speed to 0, making the speed constant.
As noted by other authors \citep[e.g.][]{styleGAN}, for complicated datasets like the Imagenet, it might be better to apply the regularizer to the feature space. 
In the experiments below, we apply the regularizer directly to the data space because the dataset is simple.

\section{Experiments}
\label{appx:exps}

\subsection{Comparing evaluation metrics}
\begin{figure*}[!ht]
\centering
\subfigure[]{
\adjincludegraphics[width=.4\textwidth, trim={{0.05\width} {0.05\width} {0.1\width} {0.12\width}}, clip]{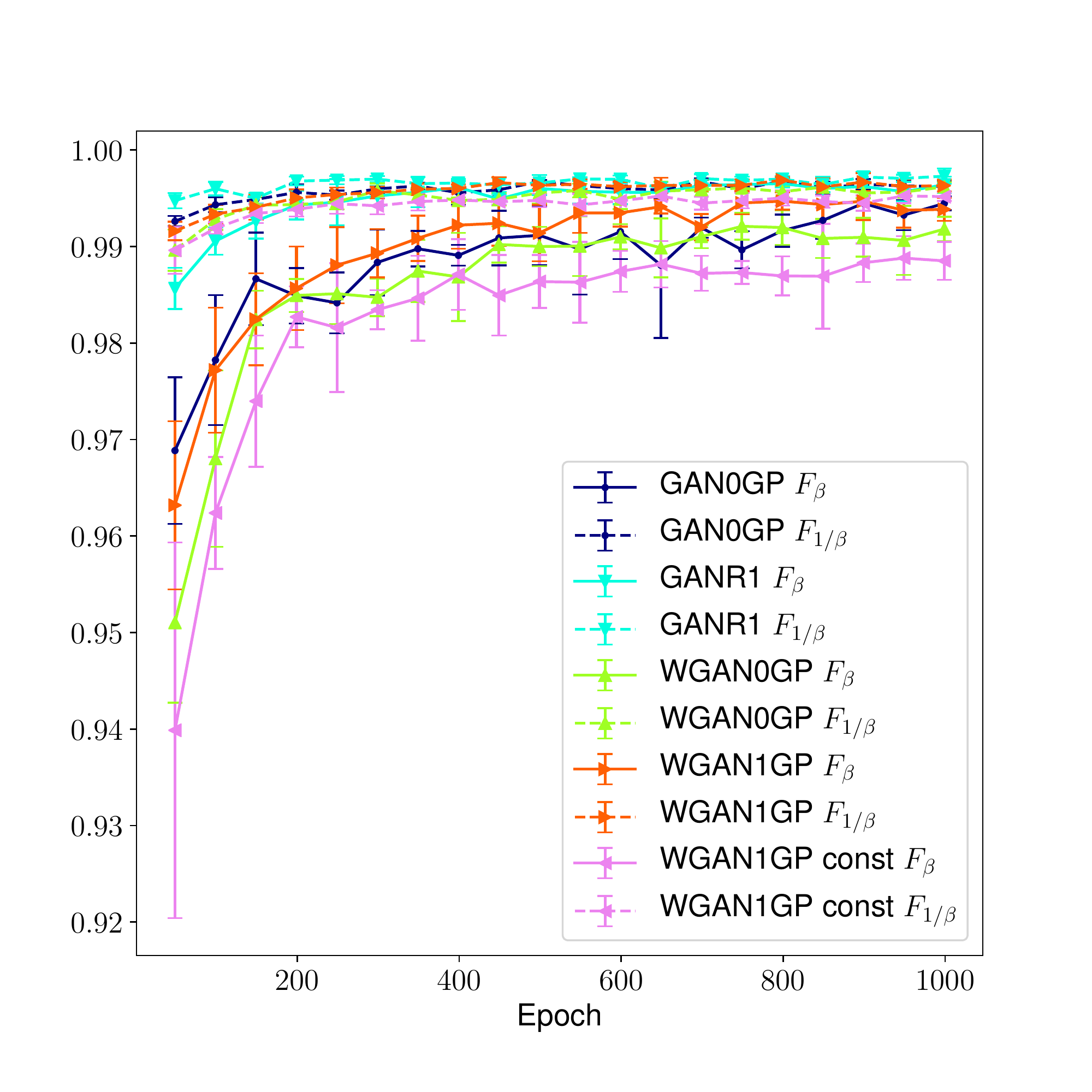}
\label{fig:kmeansWGAN}}
\subfigure[]{
\adjincludegraphics[width=.4\textwidth, trim={{0.05\width} {0.05\width} {0.1\width} {0.12\width}}, clip]{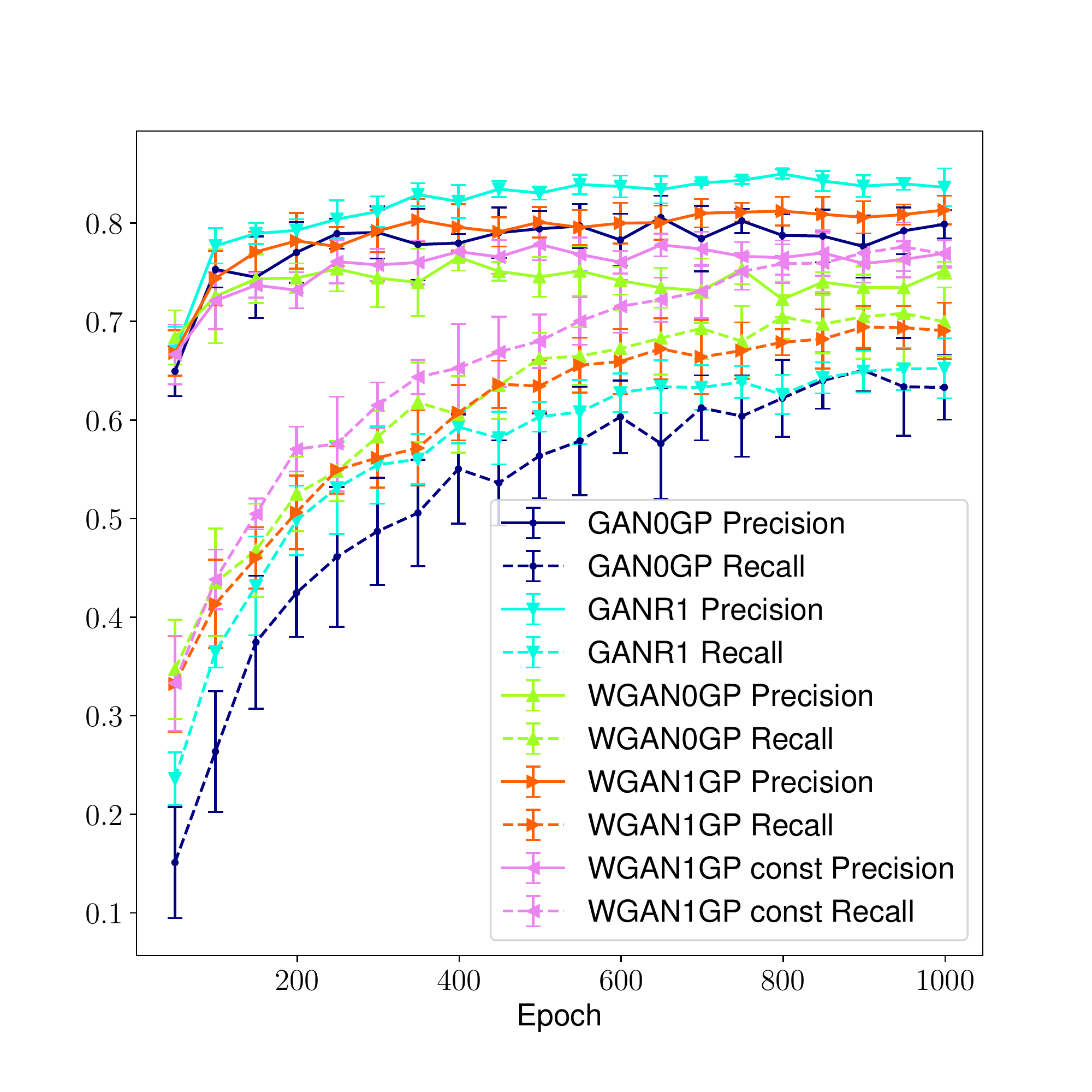}
\label{fig:knnWGAN}}
\subfigure[]{
\adjincludegraphics[width=.4\textwidth, trim={{0.05\width} {0.05\width} {0.1\width} {0.12\width}}, clip]{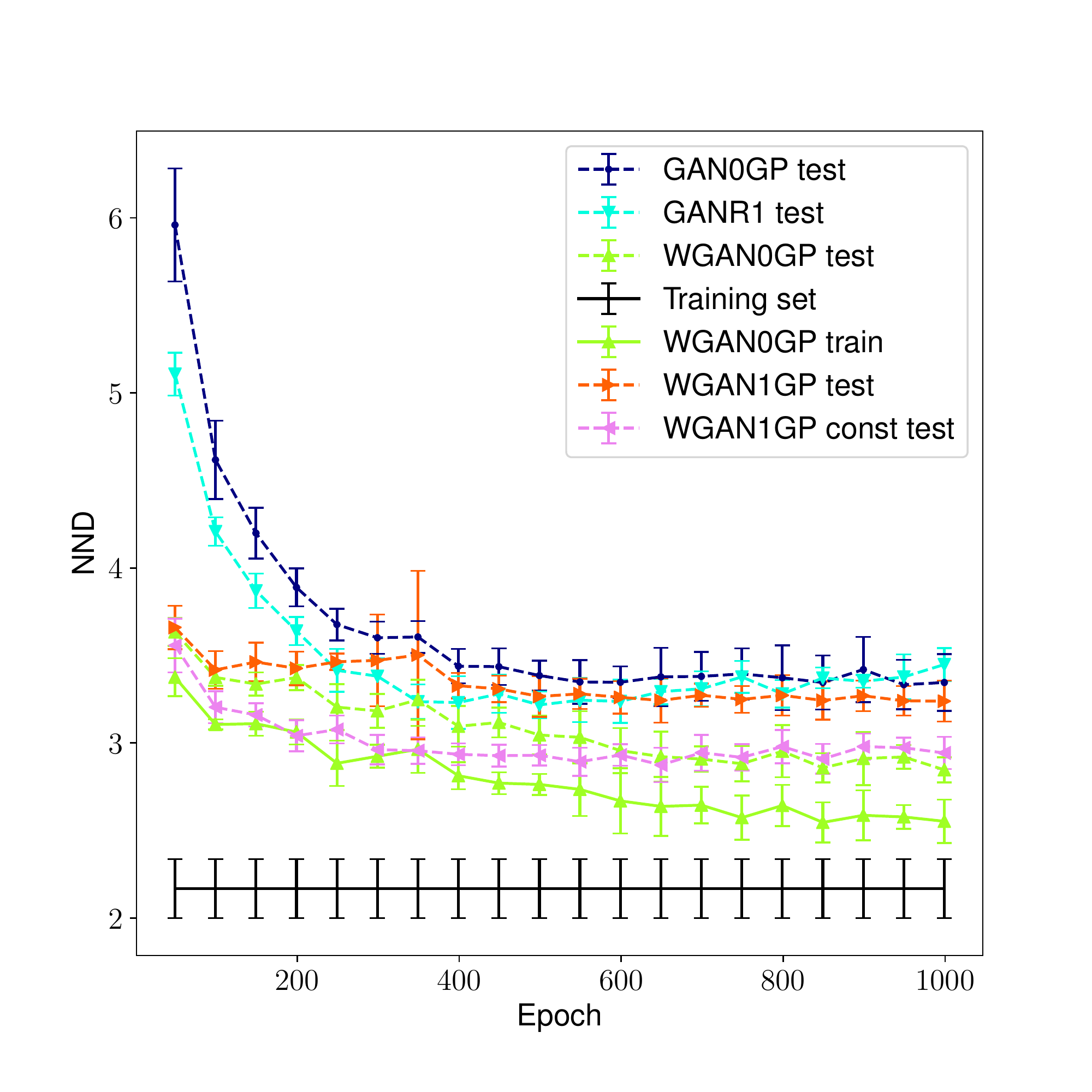}
\label{fig:nndWGAN}}
\subfigure[]{
\begin{minipage}[b]{0.4\textwidth}
\vspace{0pt}
\adjincludegraphics[width=\textwidth, trim={{0.0\width} {0.0\height} {0.0\width} {0.02\width}}, clip]{figs/gan0gp/knn_pr.pdf} 

\adjincludegraphics[width=\textwidth, trim={{0.0\width} {0.93\height} {0.75\width} {0.\width}}, clip]{figs/gan0gp/1589733702.8505816/epoch_449_inter.jpg} 

\adjincludegraphics[width=\textwidth, trim={{0.25\width} {0.93\height} {0.5\width} {0.\width}}, clip]{figs/gan0gp/1589733702.8505816/epoch_449_inter.jpg} 	

\adjincludegraphics[width=\textwidth, trim={{0.5\width} {0.93\height} {0.25\width} {0.\width}}, clip]{figs/gan0gp/1589733702.8505816/epoch_449_inter.jpg} 	
\end{minipage}
\label{fig:gan0gpBad}}
%

\caption[Evaluating evaluation metrics]{
We trained 2 sets of GANs with 2 different learning rates.
We deliberately chose a large learning rate for the second group to make it collapse.
\subref{fig:kmeansWGAN} - \subref{fig:nndWGAN} results of $k$-means PR, $k$-nn PR, and NND for models trained with $lr_1 = 2\times 10^{-4}$. 
\subref{fig:nndWGAN} Because $\fnnd{\ptrain}{\phat_{fake}}$ and $\fnnd{\ptest}{\phat_{fake}}$ always show the same trend, we show only $\fnnd{\ptest}{\phat_{fake}}$. We show $\fnnd{\ptrain}{\phat_{fake}}$ for WGAN0GP as a representative.
\subref{fig:gan0gpBad} top: $k$-nn PR of GAN0GP trained with $lr_2 = 10^{-3}$, bottom: generated samples from a GAN with recall equal 1.
}
\label{fig:result}
\end{figure*}

We measure the performance of evaluation metrics on GANs trained on MNIST to see how their worst-case performances correlate to their real-world performances.
The result is shown in Fig.~\ref{fig:kmeansWGAN} - \ref{fig:nndWGAN}.\footnote{IS and FID are not included because computing them requires the Inception model.
\citet{towardGenMetricGAN} empirically showed that IS and FID are not robust against training set memorization.
We refer readers to their experimental results.}
We trained 5 variants of GAN on the MNIST dataset. The variants are GAN with 0GP regularizer \citep{improveGeneralization} (GAN0GP), GAN with R1 regularizer \citep{whichGANConverge} (GANR1), WGAN with gradient penalty \citep{wgan, wgangp} (WGAN-1GP), WGAN with 0GP \citep{improveGeneralization}, and WGAN1GP with our new constant speed regularizer (WGAN1GP-const).
\citet{improveGeneralization} showed in their paper that 0GP improves the generalization capacity of the discriminator and that, in turn, improves the generator's generalization capacity.
They also claimed that 1GP \citep{wgangp} does not help improving generalization.
We verify their claims with our fixed NND metric.

The result for NND is shown in Fig.~\ref{fig:nndWGAN}.
WGAN0GP consistently has lower NND than WGAN1GP.
That confirms the claims by \citet{improveGeneralization}.
We also find that GAN0GP also more stable than GANR1 as GANR1 can suddenly collapse during training.
Fig.~\ref{fig:nndWGAN} shows that GANR1 suffers from overfitting after epoch 400, and its NND starts to increase.
GAN0GP's NND decreases as the training progresses.
Our constant speed does a very good job at improving the generalization capacity of WGAN1GP.
WGAN1GP-const has almost the same NND as WGAN0GP while producing smoother interpolation (see Fig.~\ref{fig:constSpeedMNIST}, \ref{fig:wgan1gpMNIST}). 

$k$-means PR does not do a good job at distinguishing models with different generalization capacities.
Fig.~\ref{fig:kmeansWGAN} shows that all models achieve almost perfect precision (i.e. $F_{1/\beta}$) after 400 epochs.
In contrast to NND, $k$-means PR assigns the highest precision and recall to GANR1.
This result is expected as \citet{catastrophicGAN} and \citet{improveGeneralization} suggested that the R1 regularizer may encourage the generator to memorize the training set.
The other GANs with better generalization capacity will produce more data points that are further away from the training and test sets.
The far data points result in a lower recall, as we see in Fig.~\ref{fig:kmeansWGAN}.

$k$-nn PR does a much better job distinguishing different models (Fig.~\ref{fig:knnWGAN}).
The result seems to justify the claim of its authors \cite{precisionRecallKnn}.
However, Fig.~\ref{fig:gan0gpBad} shows that $k$-nn PR assigns very high recall to a very bad GAN.
Fig.~\ref{fig:knn} illustrates the same phenomenon on a lower-dimensional dataset.
$k$-nn works fine for good models that distribute the fake data evenly across the real manifold, but it is vulnerable to bad models that generate many outliers.

\textbf{Discussion}: This experiment shows that the robustness of evaluation metrics on real-world datasets is directly correlated to the size of their minimal dataset $\data^*$.
$k$-nn PR has the smallest minimal set and is the most vulnerable to outliers in the data.
$k$-means PR has $\abs{\data^*_{kmeans}} = k$ and different data points in $\data^*_{kmeans}$ must lie in different clusters.
To fool $k$-means PR, a generator must produce fake data points belonging to different clusters.
That is nontrivial for real-world generative models.
NND (both the original and our updated version) has the biggest $\data^*_{nnd}$ and is the most robust.
Our updated NND can be used to estimate the generalization capacity of GMs.

\subsection{Evaluating our constant speed regularizer and MDL based metric}
\begin{figure}[!ht]
\centering
\subfigure[]{
\includegraphics[width=0.19\textwidth]{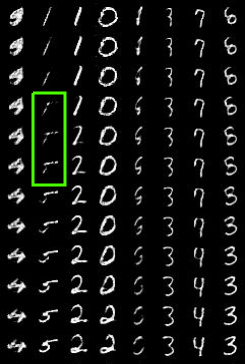} \label{fig:constSpeedMNIST}
}
\subfigure[]{
\includegraphics[width=0.19\textwidth]{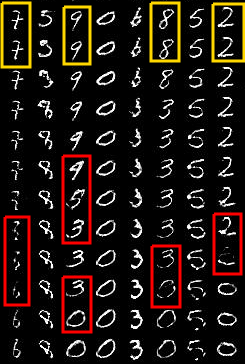} \label{fig:wgan1gpMNIST}
}
\caption[Interpolation results for WGAN1GP and WGAN1GP-const]{\subref{fig:constSpeedMNIST}, \subref{fig:wgan1gpMNIST} interpolation result of WGAN1GP-const and WGAN1GP at epoch 1000.
Despite the lower NND, WGAN0GP produces similar interpolation result as WGAN1GP.}
\label{fig:interpolate}
\end{figure}

Fig.~\ref{fig:interpolate} shows the interpolation results for WGAN1GP and WGAN1GP-const.
Interpolations produced by WGAN1GP contain many sudden jumps (red boxes in Fig.~\ref{fig:wgan1gpMNIST}).
At the beginning and the end of an interpolation path, the image at a step is almost identical to the image at the previous step (yellow boxes in Fig.~\ref{fig:wgan1gpMNIST}).
The phenomenon suggests that similar to the generator in Fig.~\ref{fig:interExplode}, WGAN1GP memorizes the training data.
In contrast, WGAN1GP-const produces smooth interpolations with no sudden jumps.
Fig.~\ref{fig:nndWGAN} shows that WGAN1GP-const has a very good generalization capacity, i.e., it can produce many more images that are not in the training set.
Our constant speed regularizer helps the generator avoids training set memorization and generalizes.

\begin{figure}[!ht]
\centering
\subfigure[]{
\adjincludegraphics[width=.5\textwidth, trim={{0.05\width} {0.05\width} {0.1\width} {0.12\width}}, clip]{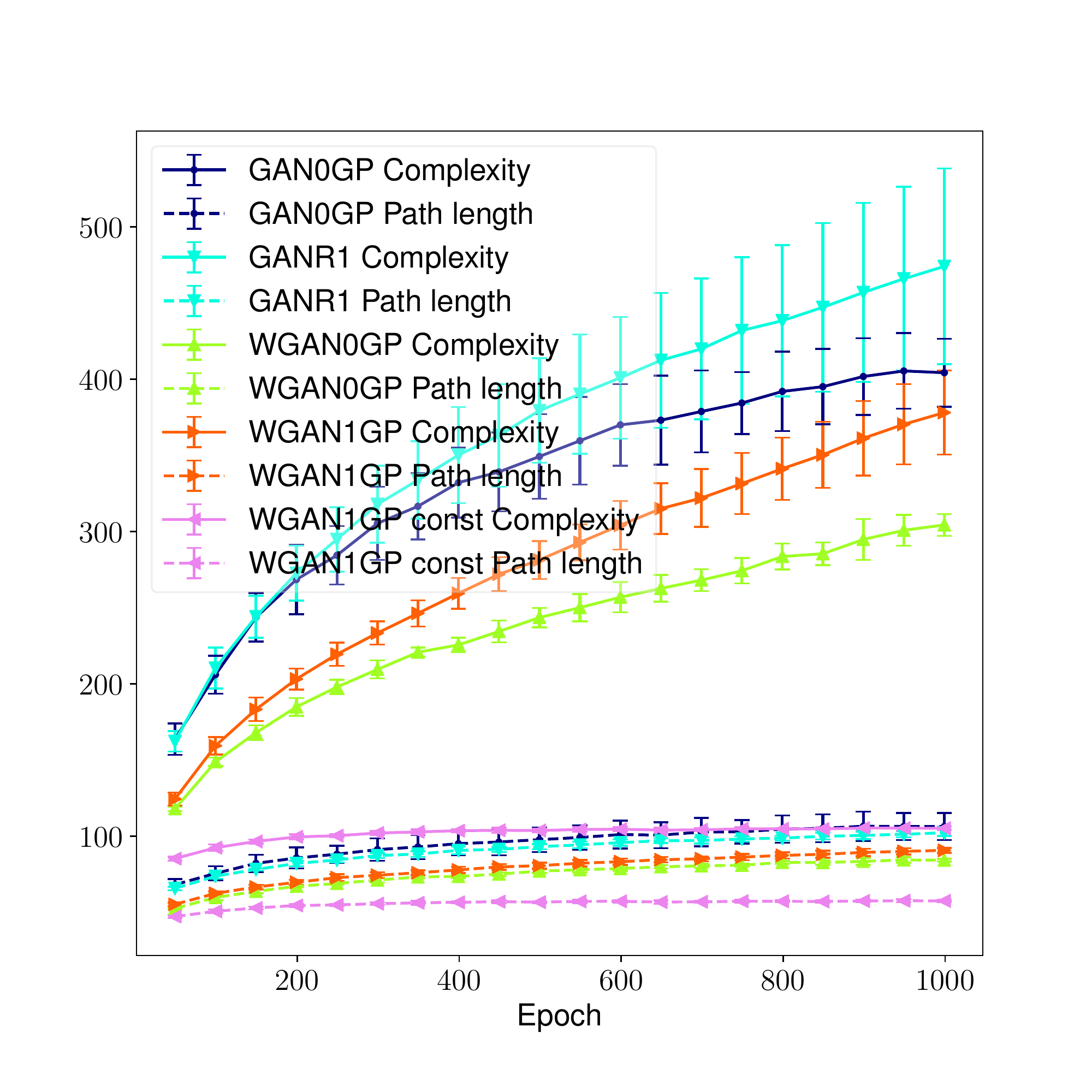}}
\caption[Experimental result for the computational complexity]{Experimental result for the computational complexity.}
\label{fig:mdl}
\end{figure}

We test the discriminative power of the computational complexity.
We compare the computational complexity with the pairwise path length \citep{styleGAN}.
The result is shown in Fig.~\ref{fig:mdl}.
The pairwise path length is not good for distinguishing different models.
WGAN1GP and WGAN0GP have roughly the same pairwise path length.
The same is true for GANR1 and GAN0GP.
In contrast, the computational complexity can clearly separate different models.
We also see that the complexity of each model increases as the training progresses.
GANs with 0GP have lower complexity than GANs with 1GP and R1 penalty.
The result confirms that 0GP can help to reduce the complexity and improve generalization.
Our constant speed regularizer does an outstanding job in minimizing the complexity.
The complexity of WGAN1GP-const is much lower than other models and does not increase as the training progresses.
The result again confirms the effectiveness of our constant speed regularizer.

\section{Limitations and Future Work}

In Section \ref{sec:capacity}, we introduce a method for comparing evaluation metrics.
To compute the capacity of an evaluation metric $f$, we have to construct its minimal dataset $\data^*_f$.
However, the minimal dataset is not always easy to construct and might depend on the evaluation metric's parameters (e.g., the NND).
We need a general method for estimating the minimal datasets of evaluation metrics.

Our generalization metric uses $\alpha$ to control the relative importance of the divergence to the complexity.
Although we discussed the effect of $\alpha$ on the metric in Section \ref{sec:metricDef}, we have not yet tested its effect in practical settings.
An algorithm for determining the optimal value for $\alpha$ is also needed.
They are open problems that we aim to solve in our next work.

\section{Conclusion}
In this paper, we investigate the ability to measure generalization of different evaluation metrics.
We show that all of the existing metrics can be fooled by training set memorization.
We study the worst-case performance of common metrics and show that their worst-case performance is related to their robustness in real-world scenarios.
We identify the weakness of NND and propose a fix to the problem.
Our experiments show that the fixed NND metric is robust to noise in the input and can be used to estimate the generalization capacities of generative models.
We introduce a constant speed regularizer for improving the smoothness of the generated data manifold and the generalization capacity of the generator in generative latent variable models.
We develop a novel metric for generalization based on the MDL principle.
Initial results show that our metric has better discriminative power than existing metrics.

\bibliographystyle{abbrvnat}
\bibliography{dgmgen}

\appendix

\section{Inception Score}
\label{appx:is}

We assume that the Inception model is a perfect classifier, i.e. given a real datapoint $\bm x$ the conditional distribution $p(y | \bm x)$ is 
\begin{eqnarray}
p(y = i | \bm x) = \begin{cases}
1 \text{ if } i \text{ is the correct label of } \bm x \\
0 \text{ otherwise}
\end{cases}
\end{eqnarray}
We have $\text{H}(y | \bm x) = 0$.
We select $C$ real datapoints from $C$ classes in the training set so $\text{H}(y) = \ln C$.
The Inception score for the dataset in Eqn.~\ref{eqn:isD} is $C$.

\section{Neural Net Divergence}
\label{appx:nnd}

The configuration for the experiment in Fig.~\ref{fig:noiseInf} and \ref{fig:noiseFixed} is given in Table \ref{tab:nndNoise}.

\begin{figure}
\centering
\subfigure[$\epsilon = 0$]{\includegraphics[width=0.3\textwidth]{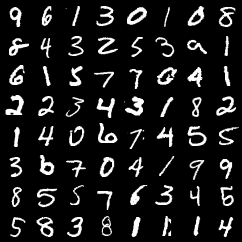}}
\subfigure[$\epsilon = 0.1$]{\includegraphics[width=0.3\textwidth]{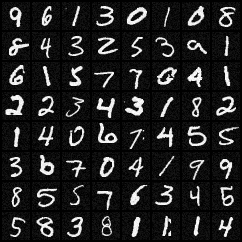}}
\subfigure[$\epsilon = 0.5$]{\includegraphics[width=0.3\textwidth]{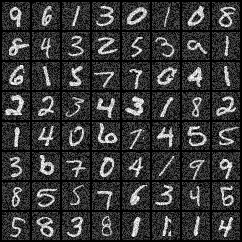}}\\
\subfigure[$\epsilon = 1$]{\includegraphics[width=0.3\textwidth]{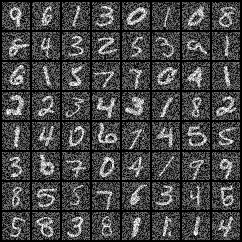}}
\subfigure[$\epsilon = 2$]{\includegraphics[width=0.3\textwidth]{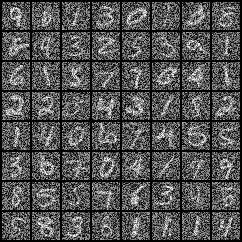}}
\subfigure[$\epsilon = 5$]{\includegraphics[width=0.3\textwidth]{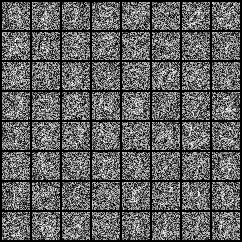}}
\caption{Noisy images generated by our procedure with $p_u = \mathcal{U}(-1, 1)$.}
\label{figx:noisyImages}
\end{figure}

\begin{table}
\centering
\caption{Configuration for experiment in Fig.~\ref{fig:noiseInf}}
\label{tab:nndNoise}
\begin{tabular}{|c|c|}
\hline
Network architecture & MLP with 3 hidden layers, 512 hidden neurons \\
Loss function & WGAN1GP loss function \cite{wgangp} \\
Number of training iteration & 20000 \\
Learning rate & 1e-4 \\
Optimizer & Adam with $\beta_1 = 0.9,\ \beta_2 = 0.999$\\
$\abs{\dtest}$ & 10000 \\
\hline
\end{tabular}
\end{table}

The GANs in Fig.~\ref{fig:infVSFixed} and \ref{fig:result} also use 3 hidden layer MLPs with the same number of hidden neurons.

\section{$k$-means Precision-Recall}
\label{appx:kmeans}

\begin{minipage}{0.5\textwidth}
\begin{algorithm}[H]
    \centering
    \caption{$k$-means PR} \label{algox:kmeans}
    \begin{algorithmic}[1]
        \State \textbf{Inputs}: $\mathcal{D}_r,\ \mathcal{D}_g$, hyper parameter $\beta$
        \State \textbf{Outputs}: $ F_\beta,\ F_{1/\beta}$
        \State $\mathcal{C} \gets kmeans(\mathcal{D}_r \cup \mathcal{D}_g)$
        \State $hist_g[i] \gets \frac{\text{\# fake datapoints in cluster } \mathcal{C}_i}{|\mathcal{D}_g|}$
        \State $hist_r[i] \gets \frac{\text{\# real datapoints in cluster } \mathcal{C}_i}{|\mathcal{D}_r|}$
        \State $curve = pr\_curve(hist_r, hist_g)$
        \State ${F_\beta = f\_score(curve, \beta)}$
        \State ${F_{1/\beta} = f\_score(curve, 1/\beta)}$
        \State \Return $ F_\beta,\ F_{1/\beta}$
    \end{algorithmic}
\end{algorithm}
\end{minipage}

We describe the procedure for constructing $\data^*_{kmeans}$.
Without loss of generality, we assume that $\abs{\dtest} = \abs{\dg}$, and the $k$-means algorithm always finds the optimal clustering.
$k$-means PR applies $k$-means clustering algorithm on $\dtest \cup \dg$ to get $k$ clusters. 
Let $\mathcal{C}_i$ be the $i$-th cluster, $\mathcal{C}_{i, test}$ be the set of test datapoints in this cluster.
From the algorithm above, we see the the perfect precision and recall are achieved when $hist_g = hist_r$.
To make $hist_g[i] = hist_r[i]$, the generator $G$ needs to
\begin{enumerate}
\item  memorize a training datapoint $\bm x$ s.t. $\bm x$ is closest to the center of cluster $\mathcal{C}_i$.
\item duplicate the datapoint $\abs{\mathcal{C}_{i, test}}$ times
\end{enumerate}

The number of distinct datapoints that $G$ has to memorize is $k$, the number of clusters.

\section{$k$-nn Precision-Recall}
\label{appx:knn}

\begin{minipage}{0.7\textwidth}
\begin{algorithm}[H]
    \centering
    \caption{$k$-nn PR} \label{algox:knn}
    \begin{algorithmic}[1]
        \State \textbf{Inputs}: $\mathcal{D}_r,\ \mathcal{D}_g$
        \State \textbf{Outputs}: Precision $P$, Recall $R$
        \State $\mathcal{M}_g \gets \emptyset$ \quad // build the fake manifold
        \For{\text{each datapoint } $\bm x_{i} \in \mathcal{D}_g$}
        \State $\bm x_j \gets $ the $k$-th nearest neighbor of $\bm x_i$
        \State $S_i \gets sphere(c=\bm x_i, r=\norm{\bm x_i - \bm x_j})$
        \State $\mathcal{M}_g \gets \mathcal{M}_g \cup S_i$
        \EndFor
        \State \text{build the real manifold } $\mathcal{M}_r$
        \State $P \gets \frac{\text{\# fake datapoints in } \mathcal{M}_r}{|\mathcal{D}_g|}$
        \State $R \gets \frac{\text{\# real datapoints in } \mathcal{M}_g}{|\mathcal{D}_r|}$
        \State \Return $P,\ R$
    \end{algorithmic}
\end{algorithm}
\end{minipage}

The idea for constructing $\data^*_{knn}$ is illustrated in Fig.~\ref{fig:knn}. 
We select $\bm x_1, \bm x_2 \in \dtrain$ s.t. $\norm{\bm x_1 - \bm x_2} \ge \norm{\bm x_i - \bm x_j} \ \forall\ \bm x_i, \bm x_j \in \dtrain$.
Because $\bm x_1, \bm x_2$ are in $\dtrain$, they lie on the real manifold.
Thus, the precision is 1.
Because $\norm{\bm x_1 - \bm x_2} \ge \norm{\bm x_i - \bm x_j} \ \forall\ \bm x_i, \bm x_j \in \dtrain$, all of the training datapoints are in the two spheres centered at $\bm x_1$ and $\bm x_2$.
If $\dtrain$ is large enough then we can assume that all of the test datapoints lies in the manifold reconstructed using $\dtrain$.
Therefore, all of the test datapoints are highly likely to lie in the two spheres centered at $\bm x_1, \bm x_2$ and the recall is 1.

%

\end{document}